\documentclass[lettersize,journal]{IEEEtran}
\usepackage{amsmath,amsfonts}
\usepackage{array}
\usepackage[caption=false,font=normalsize,labelfont=sf,textfont=sf]{subfig}
\usepackage{booktabs}
 \usepackage{algorithm}
\usepackage{algpseudocode}
\usepackage{float}
\usepackage{amsfonts}
\usepackage{multirow}
\usepackage{soul}
\usepackage{xcolor}
\usepackage[square,numbers]{natbib}
\usepackage{comment}
\usepackage{graphicx}

\usepackage{enumitem}
\setlist{nosep, leftmargin=14pt}
\usepackage{color}
\usepackage{hyperref}
\begin{document}

\title{\texttt{Cross-D Conv}: Cross-Dimensional Transferable Knowledge Base \\ via Fourier Shifting Operation}

\author{Mehmet Can Yavuz and Yang Yang\\ Department of Radiology and Biomedical Imaging\\ University of California - San Francisco, USA.\\
    \texttt{\{mehmetcan.yavuz, yang.yang4\}@ucsf.edu}}

\maketitle

\begin{abstract}
In biomedical imaging analysis, the dichotomy between 2D and 3D data presents a significant challenge. While 3D volumes offer superior real-world applicability, they are less available for each modality and not easy to train in large scale, whereas 2D samples are abundant but less comprehensive. This paper introduces \texttt{Cross-D Conv} operation, a novel approach that bridges the dimensional gap by learning the phase shifting in the Fourier domain. Our method enables seamless weight transfer between 2D and 3D convolution operations, effectively facilitating cross-dimensional learning. The proposed architecture leverages the abundance of 2D training data to enhance 3D model performance, offering a practical solution to the multimodal data scarcity challenge in 3D medical model pretraining. Experimental validation on the RadImagenet (2D) and multimodal volumetric sets demonstrates that our approach achieves comparable or superior performance in feature quality assessment. The enhanced convolution operation presents new opportunities for developing efficient classification and segmentation models in medical imaging. This work represents an advancement in cross-dimensional and multimodal medical image analysis, offering a robust framework for utilizing 2D priors in 3D model pretraining while maintaining computational efficiency of 2D training. \textit{Codes \& Weights}: \url{https://github.com/convergedmachine/Cross-D-Conv}
\end{abstract}


\section{Introduction}
\label{sec:intro}

The field of computer vision has traditionally focused on 2D image analysis due to its relative simplicity and suitability for a wide range of tasks, including those in biomedical imaging. However, the medical domain often requires the analysis of 3D images.

The recent emergence of large-scale volumetric biomedical datasets has significantly advanced medical imaging research \cite{liwell}. However, training models from scratch on these extensive volumetric datasets for subsequent fine-tuning on specific tasks can be time-consuming. Transfer learning strategies provide an effective and viable solution to this challenge, particularly when 2D pretraining proves successful for 3D applications. Transitioning from 2D to 3D analysis presents substantial challenges that require sophisticated solutions. While 2D datasets are abundant and well-curated across various modalities, their 3D counterparts remain limited in both scale and diversity. Directly adapting existing 2D datasets for 3D applications introduces significant technical hurdles, underscoring the importance of cross-dimensional learning approaches. These challenges raise a fundamental research question: Can we design a novel convolutional operation that effectively bridges the gap between cross-dimensional training paradigms?

\begin{figure}[b!]
    \begin{center}
    \includegraphics[width=\columnwidth]{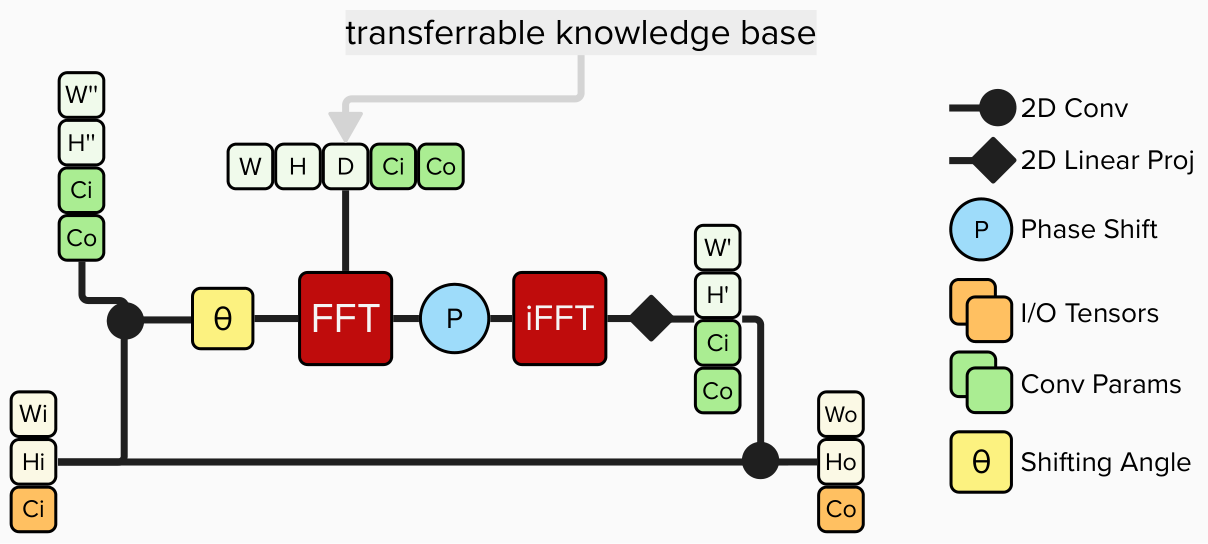}
    \end{center}
    \caption[Cross-Dimensional Convolution Operation Pipeline]{Architectural diagram of the \texttt{Cross-D Conv} operation workflow. The process transforms 2D input tensors through: (1) rotation parameter generation from spatial coordinates, (2) Fourier transform and phase shifting, and (3) projection of 3D convolutional weights onto 2D kernels. Green blocks indicate trainable parameters, while orange blocks represent I/O tensors.}
  \label{Fig:Alg}
\end{figure}

To address this fundamental research question, we investigate the intricate relationship between 2D and 3D convolutional operations within the framework of cross-dimensional learning. Our hypothesis posits that meaningful correlations exist between 2D and 3D feature representations, which can be leveraged to achieve more efficient and robust volumetric analysis. Specifically, if models trained on 2D data can demonstrably enhance performance on 3D tasks, this would indicate a substantial transferable knowledge base across dimensions. Such insights would pave the way for the development of innovative hybrid architectures capable of simultaneous 2D and 3D processing, thereby harnessing the strengths of both paradigms.

Building upon existing work, such as the enhanced convolution operation proposed \cite{Yang2021}, which adapts pre-trained 2D architectures for 3D image analysis by applying 2D convolutional filters along all three axes to create pseudo-3D kernels (ACS-Convolution), our approach similarly aims to re-utilize information learned in the 2D domain to enhance the effectiveness of 3D image analysis.

In this study, we introduce the \texttt{Cross-D Conv} operation, a novel convolutional method that bridges the dimensional gap by shifting the phases of the kernels in the Fourier domain. This technique enables seamless weight transfer between 2D and 3D convolution operations, thereby facilitating cross-dimensional learning and establishing a demonstrable correlation between 2D and 3D training processes.

Experimental validations on RadImagenet (2D) and various multimodal volumetric datasets demonstrate that our method achieves comparable or superior performance in feature quality assessment relative to conventional approaches. These results represent a significant advancement in cross-dimensional and multimodal medical image analysis, highlighting the potential of the \texttt{Cross-D Conv} operation to enhance both efficiency and effectiveness in volumetric medical imaging tasks.

\section{Methodology}

In this study, we introduce the \texttt{Cross-D Conv} module, a novel convolutional layer designed to enhance feature extraction through dynamic weight manipulation. It integrates rotation-based transformations and Fourier transformations to capture multi-dimensional complex spatial patterns effectively. Below, we detail the architecture and operational workflow of the \texttt{Cross-D Conv} module. The algorithmic operations we discuss is presented at Algorithm 1 and demonstrated in Figure \ref{Fig:Alg}.

\noindent \textbf{Transferable Knowledge Base}
At the core of the \texttt{Cross-D Conv} module lies a learnable 3D convolutional weight or 5D transferable knowledge base, denoted as $\mathbf{U}_{3D} \in \mathbb{R}^{C_{\text{out}} \times C_{\text{in}} \times K \times K \times K}$, where $C_{\text{in}}$ and $C_{\text{out}}$ are the input and output channels, and $K$ is the kernel size.

\noindent \textbf{Rotation Parameter Representation.} Let \(x \in \mathbb{R}^{C \times H \times W}\) be the input to the secondary network, and let \(f(x) \in \mathbb{R}^{4 \times (HW)}\) be the corresponding feature map it produces. We aggregate these features into a single 4D rotation parameter vector \(r\) using a softmax-weighted sum:
\begin{equation}
r = \sum_{hw=1}^{HW} f_{hw}(x) \,\cdot\, \mathrm{softmax}\bigl(f(x)\bigr)_{hw},
\end{equation}

\noindent where \(r = (k_x, k_y, k_z, \theta)\). We interpret the first three components \((k_x, k_y, k_z)\) as the rotation axis \(\mathbf{k}\), and the fourth component \(\theta\) as the rotation angle.

\noindent To ensure a valid rotation, we normalize the axis so that \(\|\mathbf{k}\| = 1\). Additionally, we constrain \(\theta\) to lie within the interval \(\left[-\frac{\pi}{4}, \frac{\pi}{4}\right]\), limiting the rotation to at most \(45^\circ\) in either direction.

\noindent \textbf{Approximate Rotation Matrix Construction} The rotation matrix $\mathbf{R}(\theta)$ is constructed using Rodrigues' formula and linear approximation for small angles:
\begin{equation}
    \mathbf{R}(\theta) \approx \mathbf{I} + \theta \mathbf{K},
\end{equation}
where 
\begin{equation}
    \mathbf{I} =
    \begin{bmatrix}
    1 & 0 & 0 \\
    0 & 1 & 0 \\
    0 & 0 & 1
    \end{bmatrix}, \quad
    \mathbf{K} =
    \begin{bmatrix}
    0 & -k_z & k_y \\
    k_z & 0 & -k_x \\
    -k_y & k_x & 0
    \end{bmatrix}.
\end{equation}
Here, \(\mathbf{k} = (k_x, k_y, k_z)\) is the rotation axis, and \(\theta\) is the rotation angle.

\noindent \textbf{Frequency Domain Rotation via FFT}
To perform rotation efficiently, the weights are rotated in the frequency domain:
\begin{enumerate}
    \item \textbf{FFT of Convolutional Weights:} Compute the FFT of $\mathbf{U}_{3D}$:
    \begin{equation}
        \mathcal{F}\{U\} = \text{FFT}(U).
    \end{equation}
    \item \textbf{Frequency Grid Construction:} Generate frequency grids $(f_x, f_y, f_z)$ for the kernel dimensions.
    \item \textbf{Rotation of Frequency Coordinates:} Use the approximated rotation matrix \(\mathbf{R}(\theta)\) to rotate the frequency grids:
    \begin{equation}
    \begin{bmatrix}
    f'_x \\
    f'_y \\
    f'_z
    \end{bmatrix}
    =
    \mathbf{R}(\theta)
    \begin{bmatrix}
    f_x \\
    f_y \\
    f_z
    \end{bmatrix}.
    \end{equation}
    \item \textbf{Phase Shift Application:} Apply a phase shift based on rotated frequencies:
    \begin{equation}
        \Phi = \exp\bigl(-2j\pi (f'_x + f'_y + f'_z)\bigr),
    \end{equation}
    \begin{equation}
        \mathcal{F}\{U'\} = \mathcal{F}\{U\} \cdot \Phi.
    \end{equation}    
    \item \textbf{Inverse FFT:} Perform inverse FFT to compute rotated weights:
    \begin{equation}
        U' = \text{iFFT}\bigl(\mathcal{F}\{U'\}\bigr).
    \end{equation}
\end{enumerate}

\noindent \textbf{Final Weight Application} The rotated weights are used for convolution operation, leveraging the dynamically adjusted kernels.

\begin{algorithm}[b!]
\caption{\texttt{Cross-D Conv} 2D-Forward Pass}
\begin{algorithmic}[1]
\State \textbf{Input:} Input feature map \(x \in \mathbb{R}^{B \times C_{\text{in}} \times H \times W}\), 3D kernel weights \(U_{3D} \in \mathbb{R}^{C_{\text{out}} \times \frac{C_{\text{in}}}{G} \times K \times K \times K}\)
\State \textbf{Output:} Output feature map \(y \in \mathbb{R}^{B \times C_{\text{out}} \times H' \times W'}\)

\State \textbf{Step 1: Compute Rotation Parameters:}
\State \quad \((\mathbf{k}, \theta) \gets \text{f}(x)\) \Comment{Predict rotation axis and angle.}

\State \textbf{Step 2: Rotate 3D Weights:}
\State \quad \textbf{FFT of 3D Weights:}
\State \quad \quad \(U_{FFT} \gets \text{FFT}(U_{3D})\)
\State \quad \textbf{Compute Rotated Frequencies:}
\State \quad \quad Generate frequency grids \((f_x, f_y, f_z)\)
\State \quad \quad \( (f'_x, f'_y, f'_z) \gets \mathbf{R}(\theta) (f_x, f_y, f_z)\)
\State \quad \textbf{Phase Shift Application:}
\State \quad \quad \(\Phi \gets \exp\bigl(-2j\pi (f'_x + f'_y + f'_z)\bigr)\)
\State \quad \textbf{Apply Phase Shift and Inverse FFT:}
\State \quad \quad \(U'_{FFT} \gets U_{FFT} \cdot \Phi\)
\State \quad \quad \(U' \gets \text{iFFT}(U'_{FFT})\)

\State \textbf{Step 3: Extract 2D Kernels \& perform 2D Conv:}
\State \quad \(U_{2D} \gets \text{mid}(U')\) \Comment{Extract the middle slice.}
\State \quad \(y \gets \text{conv2d}(x, U_{2D})\)

\State \Return \(y\) \(\in\) \(\mathbb{R}^{B \times C_{\text{out}} \times H' \times W'}\)
\end{algorithmic}
\end{algorithm}

\begin{table*}[htb!]
    \centering
    \resizebox{0.85\textwidth}{!}{%
    \begin{tabular}{l|lcccc|c}
        \toprule
        Dataset & Model (ResNet18) & Precision (Macro) & Recall (Macro) & F1 (Macro) & Balanced Accuracy & Average Accuracy \\
        \midrule
        IN1K & Regular & 0.6807 & 0.6693 & 0.6657 & 0.6693 & 0.6693 \\ 
        & Cross-D Conv & \textbf{0.6895} & \textbf{0.6881} & \textbf{0.6838} & \textbf{0.6881} & \textbf{0.6881} $\uparrow$ \\        
        \midrule
        RIN & Regular & 0.5830 & 0.4989 & 0.5252 & 0.4989 & 0.8305 \\ 
        & Cross-D Conv & \textbf{0.5891} & \textbf{0.5228} & \textbf{0.5471} & \textbf{0.5228} & \textbf{0.8374} $\uparrow$ \\
        \bottomrule
    \end{tabular}
    }
    \caption{Performance Comparison of Regular and Cross-D Conv Models Using ResNet18 on Imagenet and RadImagenet Datasets. The table presents macro-averaged Precision, Recall, F1 score, Balanced Accuracy, and Average Accuracy for both Regular and Cross-D Conv variants of the ResNet18 model. \textbf{Bold} indicates higher average accuracy and up arrow ($\uparrow$) is the statistically significant improvement.}
    \label{tab:radimagenet}
\end{table*}

\begin{table*}[htb!]
\centering
\small
\resizebox{\textwidth}{!}{%
\begin{tabular}{l|lccccccccccccc}
\toprule
\multicolumn{2}{l}{Image Datasets} & \multicolumn{2}{c}{OrganC \cite{xu2019efficient} (CT)} & \multicolumn{2}{c}{OrganS \cite{bilic2023liver} (CT)} & \multicolumn{2}{c}{Brain Tumor \cite{cheng2015enhanced} (MRI)} & \multicolumn{2}{c}{Brain Dataset \cite{yavuz2025policy} (MRI)} & \multicolumn{2}{c}{Breast \cite{al2020dataset} (US)} & \multicolumn{2}{c}{Breast Cancer \cite{gomez2024bus} (US)} & \multicolumn{1}{c}{} \\
\cmidrule(lr){1-2} \cmidrule(lr){3-4} \cmidrule(lr){5-6} \cmidrule(lr){7-8} \cmidrule(lr){9-10} \cmidrule(lr){11-12} \cmidrule(lr){13-14} 
Dataset & Method & mean & std & mean & std & mean & std & mean & std & mean & std & mean & std & Average \\
\toprule
IN1K & 2D Conv & 0.862 & 0.006 & 0.708 & 0.035 & 0.884 & 0.011 & 0.305 & 0.023 & 0.819 & 0.019 & 0.745 & 0.024 & 0.720 \\
 & Cross-D Conv & \underline{0.871} & 0.007 & \underline{0.763} & 0.008 & \underline{0.892} & 0.010 & \underline{0.308} & 0.026 & \underline{0.836} & 0.021 & \underline{0.759} & 0.022 & \textbf{0.738} $\uparrow$ \\ \midrule
RIN & 2D Conv & 0.842 & 0.006 & 0.742 & 0.008 & 0.902 & 0.010 & 0.268 & 0.023 & 0.832 & 0.021 & 0.762 & 0.016 & 0.725 \\
 & Cross-D Conv & \underline{0.848} & 0.008 & \underline{0.743} & 0.008 & \underline{0.910} & 0.013 & \underline{0.283} & 0.023 & \underline{0.835} & 0.037 & \underline{0.747} & 0.024 & \textbf{0.728} \\ \bottomrule
\end{tabular}
}

\resizebox{\textwidth}{!}{%
\begin{tabular}{l|lccccccccccccccc}
\toprule
\multicolumn{2}{l}{Volumetric Datasets} & \multicolumn{2}{c}{Mosmed \cite{morozov2020mosmeddata} (CT)} & \multicolumn{2}{c}{Lung Aden. \cite{feng2020ct} (CT)} & \multicolumn{2}{c}{Fracture \cite{jin2020deep} (CT)} & \multicolumn{2}{c}{BraTS21 \cite{labella2023asnr} (MRI)} & \multicolumn{2}{c}{IXI \cite{IXI_dataset} (MRI)} & \multicolumn{2}{c}{BUSV \cite{lin2022new} (US)} & \multicolumn{1}{c}{} \\
\cmidrule(lr){1-2} \cmidrule(lr){3-4} \cmidrule(lr){5-6} \cmidrule(lr){7-8} \cmidrule(lr){9-10} \cmidrule(lr){11-12} \cmidrule(lr){13-14}
Dataset & Method & mean & std & mean & std & mean & std & mean & std & mean & std & mean & std & Average \\
\toprule
IN1K & ACS-Conv & \underline{0.523} & 0.057 & \underline{0.532} & 0.034 & 0.456 & 0.027 & 0.539 & 0.030 & 0.542 & 0.044 & 0.559 & 0.079 & 0.525\\
 & Cross-D Conv & 0.505 & 0.068 & 0.513 & 0.071 & \underline{0.469} & 0.027 & \underline{0.549} & 0.031 & \underline{0.583} & 0.059 & \underline{0.590} & 0.064 & \textbf{0.535} $\uparrow$ \\ \midrule
RIN & ACS-Conv & 0.547 & 0.072 & \underline{0.548} & 0.034 & 0.471 & 0.034 & 0.545 & 0.041 & 0.555 & 0.046 & \underline{0.604} & 0.063 & 0.545\\
 & Cross-D Conv & \underline{0.557} & 0.102 & 0.529 & 0.058 & \underline{0.491} & 0.032 & \underline{0.558} & 0.044 & \underline{0.559} & 0.050 & 0.602 & 0.066 & \textbf{0.549} \\ \midrule
\end{tabular}%
}

\resizebox{\textwidth}{!}{%
\begin{tabular}{l|lccccccc|lccccc}
\toprule
\multicolumn{2}{l}{Other Image Datasets} & \multicolumn{2}{c}{Blood \cite{acevedo2020dataset} (Microsc.)} & \multicolumn{2}{c}{Pneumonia \cite{kermany2018identifying} (XR)} & \multicolumn{2}{c}{Derma \cite{tschandl2018ham10000} (DS)} & \multicolumn{1}{c}{} & Other Vol. Datasets & \multicolumn{2}{c}{Vessel \cite{yang2020intra} (MRA)} & \multicolumn{2}{c}{Synapse (Microsc.)} & \multicolumn{1}{c}{} \\
\cmidrule(lr){1-2} \cmidrule(lr){3-4} \cmidrule(lr){5-6} \cmidrule(lr){7-8} \cmidrule(lr){10-10} \cmidrule(lr){11-12} \cmidrule(lr){13-14} 
Dataset & Method & mean & std & mean & std & mean & std & Average & Method & mean & std & mean & std & Average \\
\toprule
IN1K & 2D Conv & 0.951 & 0.004 & 0.948 & 0.006 & 0.765 & 0.009 & 0.888 & ACS-Conv & 0.874 &0.026 &\underline{0.736} &0.017 & 0.805 \\
 & Cross-D Conv & \underline{0.953} & 0.003 & \underline{0.957} & 0.005 & \underline{0.771} & 0.009 & \textbf{0.894} $\uparrow$ & Cross-D Conv & \underline{0.887} &0.018 &0.731 &0.015 &\textbf{0.809}\\ \midrule
RIN & 2D Conv & \underline{0.923} & 0.004 & 0.945 & 0.006 & 0.724 & 0.008 & \textbf{0.864} & ACS-Conv & 0.886 &0.018 &0.727 &0.013 & 0.806 \\
 & Cross-D Conv & 0.909 & 0.005 & \underline{0.946} & 0.008 & \underline{0.726} & 0.009 & 0.860 & Cross-D Conv & \underline{0.887} &0.018 &\underline{0.731} &0.015 &\textbf{0.809}\\ \bottomrule
\end{tabular}
}

\caption{Performance Comparison of 2D Conv, ACS-Conv, and Cross-D Conv Methods Across Diverse Image and Volumetric Datasets. This table presents the mean and standard deviation of performance metrics for various convolutional approaches applied to a wide range of image-based and volumetric datasets. \textbf{Bold} indicates higher average accuracy and up arrow ($\uparrow$) is the statistically significant improvement.}
\label{tab:weakprobe}
\end{table*}

\begin{table}[h!]
\centering
\setlength{\tabcolsep}{10pt}
\renewcommand{\arraystretch}{1.2}
\small
\resizebox{\linewidth}{!}{%
\begin{tabular}{lcccc}
\hline
\textbf{Shape}                & \textbf{Conv2D (ms)} & \textbf{CrossDConv (ms)} & \textbf{ACSConv (ms)} & \textbf{Conv3D (ms)} \\ \hline
(8, 64, 128, 128)             & 3.576               & 4.549                    & 10.007 & 16.487                         \\
(16, 128, 224, 224)           & 31.069              & 39.772                   & 87.332 & 199.142                        \\
(32, 384, 224, 224)           & 142.513             & 655.812                  & 663.828 & 1146.796                       \\ \hline
\end{tabular}
}
\caption{Comparison of convolution operation runtimes across different input shapes and methods.}
\label{tab:benchmark_comparison}
\end{table}

\section{Experiments}
\label{sec:experiments}
\textbf{Datasets} For pretraining, we utilized the ImageNet dataset, which comprises 1,000 classes of natural images, alongside its biomedical counterpart, RadImageNet which encompasses approximately 1.35 million annotated medical images across CT, MRI, and ultrasound modalities, categorized into 165 classes spanning 11 anatomical regions.

For finetuning, we expanded the MedMNIST dataset collection by curating additional datasets and excluding certain subsets within the CT, MRI, and ultrasound modalities. MedMNIST is a standardized collection of biomedical images designed for classification tasks on 2D and 3D images.

\textbf{Baseline Model} We selected the ResNet-18 architecture due to its modern features, such as skip connections, varying kernel sizes, and batch normalization. But typically, \texttt{Cross-D Conv} can replace any 2D convolution operation. \textit{ResNet-18 + 3D}: This is a 3D extension of the ResNet-18 architecture that can process volumetric data directly. \textit{ResNet-18 + ACS Conv} \cite{Yang2021}: Axial-Coronal-Sagittal Convolutions split 2D kernels into three parts along the channel dimension and apply them separately to the three orthogonal views of a 3D volume. This approach allows the use of pre-trained 2D weights while enabling 3D representation learning. ACS-Conv networks can be initialized with both ImageNet and RadImageNet \cite{mei2022radimagenet} priors.

\textbf{Classification Implementation Details} The training pipeline for the ResNet-18 model is configured with specific hyperparameters to optimize performance. Each of the 8xV100 GPUs processes a batch size of 32, resulting in an effective batch size of 256. We trained for 90 epochs using the AdamW optimizer with a learning rate of 0.001 and standard Torch configuration. A step learning rate scheduler reduces the learning rate by a factor of 1/10 every 30 epochs, facilitating efficient learning.

\textbf{Transfer Learning Evaluation} The weak-probing approach selectively trains only the batch normalization and linear layers while keeping other parameters frozen. This method demonstrates comparable performance in convolutional kernel quality. This focused training strategy helps maintain model efficiency while adapting to the target domain.
\section{Results and Discussion}

\textbf{ImageNet and RadImageNet Performance} Table~\ref{tab:radimagenet} compares Regular and Cross-D Conv models based on ResNet18 on ImageNet (IN1K) and RadImageNet \cite{mei2022radimagenet} (RIN). Metrics include macro-Precision, Recall, F1 score, Balanced Accuracy, and Average Accuracy.

On IN1K, Cross-D Conv outperforms Regular across all metrics: Precision increases from 0.6807 to 0.6895, Recall from 0.6693 to 0.6881, and F1 score from 0.6657 to 0.6838. Both Balanced and Average Accuracy rise to 0.6881, with Average Accuracy showing a significant improvement (0.6881).

Similarly, on RIN, Cross-D Conv performs better. Precision slightly increases from 0.5830 to 0.5891, Recall from 0.4989 to 0.5228, and F1 score from 0.5252 to 0.5471. Balanced Accuracy also improves, and Average Accuracy rises significantly from 0.8305 to 0.8374 . These results demonstrate Cross-D Conv’s consistent enhancement on both natural and medical imaging datasets.

\begin{table*}[htb!]
    \centering
    \caption{Overview of 2D Datasets Used for Fine-Tuning Evaluation}
    \label{tab:2d_datasets}
    \resizebox{\linewidth}{!}{%
    \begin{tabular}{lcccccccc}
        \toprule
        \textbf{File} & \textbf{Modality} & \textbf{Number of Samples} & \textbf{Image Dimensions} & \textbf{Pixel Range} & \textbf{Unique Labels} & \textbf{Label Type} \\
        \midrule
        Blood \cite{acevedo2020dataset}& Microscope & 17,092 & (224, 224, 3) & 0 -- 255 & 8 & Multi-class \\
        Brain \cite{yavuz2025policy} & MRI & 1,600 & (224, 224, 3) & 0 -- 255 & 23 & Multi-class \\
        Brain Tumor \cite{cheng2015enhanced} & MRI & 3,064 & (224, 224, 3) & 0 -- 255 & 3 & Multi-class \\
        Breast Cancer \cite{gomez2024bus}& US & 1,875 & (224, 224, 3) & 0 -- 255 & 2 & Binary \\
        Breast \cite{al2020dataset} & US & 780 & (224, 224, 1) & 0 -- 255 & 2 & Binary \\
        Derma \cite{tschandl2018ham10000} & Dermatology & 10,015 & (224, 224, 3) & 0 -- 255 & 7 & Multi-class \\
        OrganC \cite{xu2019efficient} & CT & 23,582 & (224, 224, 1) & 0 -- 255 & 11 & Multi-class \\
        OrganS \cite{bilic2023liver} & CT & 25,211 & (224, 224, 1) & 0 -- 255 & 11 & Multi-class \\
        Pneumonia \cite{kermany2018identifying} & XR & 5,856 & (224, 224, 1) & 0 -- 255 & 2 & Binary \\
        \bottomrule
    \end{tabular}
    }
\end{table*}

\begin{table*}[htb!]
    \centering
    \caption{Overview of 3D Datasets Used for Fine-Tuning Evaluation}
    \label{tab:3d_datasets}
    \resizebox{\linewidth}{!}{%
    \begin{tabular}{lccccccc}
        \toprule
        \textbf{File} & \textbf{Modality} &  \textbf{Number of Samples} & \textbf{Image Dimensions} & \textbf{Pixel Range} & \textbf{Unique Labels} & \textbf{Label Type} \\
        \midrule
        BraTS21 \cite{labella2023asnr}& MRI & 585 & (3, 96, 96, 96) & 0 -- 22,016 & 2 & Binary \\
        BUSV \cite{lin2022new}& US & 186 & (1, 96, 96, 96) & 0 -- 255 & 2 & Binary \\
        Fracture \cite{jin2020deep}& CT & 1,370 & (1, 64, 64, 64) & 0 -- 255 & 3 & Multi-class \\
        Lung Adenocarcinoma \cite{feng2020ct}& CT & 1,050 & (1, 128, 128, 128) & -1,450 -- 3,879& 3 & Multi-class \\
        Mosmed \cite{morozov2020mosmeddata}& CT & 200 & (1, 128, 128, 64) & 0 -- 1 & 2 & Binary \\
        Synapse \cite{yang2020intra} & Microscope& 1,759 & (1, 64, 64, 64) & 0 -- 255 & 2 & Binary \\
        Vessel \cite{medmnistv2} & MRA & 1,908 & (1, 64, 64, 64) & 0 -- 255 & 2 & Binary \\
        IXI (Gender) \cite{IXI_dataset} & MRI & 561 & (2, 160, 192, 224) & 0 -- 255 & 2 & Binary \\
        \bottomrule
    \end{tabular}
    }
\end{table*}

\href{https://huggingface.co/datasets/convergedmachine/Enhanced-MedMNIST}{\textbf{Transfer Learning Evaluation}} Table~\ref{tab:weakprobe} compares 2D Conv, ACS-Conv, and Cross-D Conv across various image and volumetric datasets, including OrganC (CT), Brain Tumor (MRI), Breast Cancer (US), and others. The table is categorized into Image and Vols based on their \href{https://arxiv.org/abs/2411.02441}{respective statistics and visuals}.

\textit{Image Datasets} Cross-D Conv consistently outperforms 2D Conv and ACS-Conv. For instance, it achieves mean accuracies of 0.871 on OrganC (CT) and 0.763 on OrganS (CT). Similar improvements are observed in Brain Tumor (MRI) and Breast Cancer (US), leading to an overall Average Accuracy increase to \textit{0.738} for IN1K and 0.728 for RIN.

\textit{Volumetric Datasets} Cross-D Conv also excels in volumetric datasets, notably achieving 0.583 on IXI (MRI) and 0.590 on BUSV (US), resulting in Average Accuracies of 0.535 for IN1K and 0.549 for RIN. This highlights its effectiveness in handling complex spatial dependencies in volumetric data.

\textit{Other Image Datasets} In other image datasets, Cross-D Conv leads in Blood (Microsc.), Pneumonia (XR), and Derma (DS) for IN1K, with Average Accuracy rising to 0.894. For RIN, it matches or slightly outperforms other methods, achieving an Average Accuracy of 0.809. These results showcase Cross-D Conv’s versatility across diverse imaging modalities.

\textit{Convolutional Operation Runtimes:} Table~\ref{tab:benchmark_comparison} compares the execution times of Conv2D, CrossDConv, ACSConv, and Conv3D across various input shapes. The results indicate that Conv2D consistently achieves the fastest runtimes, closely followed by CrossDConv, which performs significantly faster than ACSConv and Conv3D.

\textbf{Overall Discussion} The comprehensive evaluation highlights Cross-D Conv’s consistent superiority over traditional convolutional methods across various datasets and metrics. Significant improvements in Average Accuracy validate its effectiveness, positioning Cross-D Conv as a promising approach for pretraining purposes.

\section{Conclusion} This study demonstrates that pretraining on 2D data provides a valuable knowledge base for 3D tasks by linking 2D and 3D feature representations. The \texttt{Cross-D Conv} operation bridges the dimensional gap by shifting kernel phases in the Fourier domain, enabling weight transfer between 2D and 3D convolutions. Experiments on Imagenet and RadImageNet (2D) and transfer learning datasets show that \texttt{Cross-D Conv} matches or outperforms conventional methods in feature quality. Overall, 2D pretraining proves to be a robust foundation for advancing 3D medical image analysis, enhancing both accuracy and computational efficiency.

\textbf{Future Work} Future research will focus on advancing the capabilities of \texttt{Cross-D Conv} through several promising directions. \textit{Hybrid Training:} We intend to implement hybrid training methods that enable models to leverage both 2D and 3D data simultaneously, enhancing their flexibility and robustness. \textit{Optimized Classifier and Segmentation Networks:} Our efforts will also include designing specialized classifiers and segmentation networks tailored to fully exploit the potential of Cross-D Convolution, aiming for superior performance in both accuracy and efficiency. These developments will further expand the applicability and impact of \texttt{Cross-D Conv} in diverse domains.

{\appendix[Cross-Dimensional Evaluation Datasets]

Transfer learning in machine learning models, especially deep learning architectures, requires a diverse set of datasets to ensure robustness and generalizability across various tasks and domains. This appendix provides comprehensive details of the datasets employed in our evaluation, categorized into 2D and 3D datasets. The diversity in these datasets encompasses variations in image dimensions, pixel ranges, label types, and the number of unique labels, facilitating a thorough assessment of the fine-tuning capabilities.

\subsection{2D Datasets}

The 2D datasets selected for this study span a range of medical imaging modalities and classification tasks. These datasets vary in complexity, from binary classification problems to multi-class scenarios with numerous unique labels. The image dimensions are standardized to facilitate consistent processing, and the pixel ranges are uniformly scaled between 0 and 255, ensuring compatibility with common preprocessing pipelines.

\textit{Insights into 2D Datasets}

The 2D datasets encompass a variety of medical imaging modalities and classification tasks, each contributing uniquely to the evaluation process:

\textbf{Blood:} This dataset comprises 17,092 microscope images of blood samples, categorized into 8 distinct classes. The substantial number of samples and multiple classes make it suitable for assessing models' capabilities in handling complex multi-class classification tasks.

\textbf{Brain:} Consisting of 1,600 MRI images with 23 unique labels, this dataset presents a challenging multi-class classification scenario. The high number of classes allows for evaluating the model's performance in distinguishing between subtle differences in brain imaging.

\textbf{Brain Tumor:} With 3,064 MRI images divided into 3 classes, this dataset focuses on classification of brain tumors. It is instrumental in assessing the model's sensitivity and specificity in medical diagnosis applications.

\textbf{Breast Cancer:} This dataset includes 1,875 ultrasound images labeled into 2 classes (benign and malignant). It serves as a benchmark for binary classification tasks in medical imaging, particularly in cancer detection.

\textbf{Breast:} Comprising 780 ultrasound images with binary labels, this dataset, though smaller in size, is valuable for evaluating model performance in data-scarce environments.

\textbf{Derma:} Featuring 10,015 dermatological images across 7 classes, this dataset allows for testing models on skin lesion classification, a critical area in dermatology.

\textbf{OrganC:} This dataset contains 23,582 CT images with 11 unique labels, facilitating the evaluation of models in organ classification tasks using CT scans.

\textbf{OrganS:} With 25,211 CT images labeled into 11 classes, this dataset is similar to OrganC.

\textbf{Pneumonia:} Comprising 5,856 X-ray images labeled as pneumonia or normal, this dataset is essential for evaluating models in detecting lung infections, a common and critical application in medical imaging.

The standardized image dimensions (224x224 pixels) and pixel ranges (0 to 255) across these datasets ensure uniformity in preprocessing, allowing for fair comparisons of model performance. The diversity in the number of samples and unique labels provides a comprehensive platform to assess models across various complexities and data availability scenarios.

\subsection{3D Datasets}

3D datasets introduce an additional layer of complexity due to their volumetric nature, making them essential for evaluating models' capabilities in handling spatial information across three dimensions. These datasets are predominantly used in medical imaging applications such as MRI and CT scans, where volumetric data is critical for accurate diagnosis and analysis.

\textit{Insights into 3D Datasets}

The 3D datasets introduce volumetric data, adding complexity to the evaluation process and testing the models' capabilities in handling spatial information:

\textbf{BraTS21:} This dataset includes 585 MRI scans with binary labels indicating the presence or absence of brain tumors. The volumetric nature of the data (96x96x96 voxels) challenges models to accurately classify tumor regions.

\textbf{BUSV:} Comprising 186 ultrasound volumes with binary labels, this dataset focuses on breast ultrasound volumetric data, testing models' abilities in 3D medical image analysis.

\textbf{Fracture:} This dataset contains 1,370 CT volumes labeled into 3 classes, facilitating the evaluation of models in detecting and classifying bone fractures.

\textbf{Lung Adenocarcinoma:} With 1,050 CT volumes and 3 unique labels, this dataset is used to assess models in classifying different subtypes of lung adenocarcinoma, a common type of lung cancer.

\textbf{Mosmed:} This dataset includes 200 CT volumes with binary labels, focusing on the detection of COVID-19 related lung infections, providing a timely evaluation of models in pandemic-related applications.

\textbf{Synapse:} Comprising 1,759 microscope volumes with binary labels, this dataset is used for evaluating models in classifying synapses in neural imaging.

\textbf{Vessel:} This dataset contains 1,908 MRA volumes with binary labels, focusing on vessel classification tasks, essential for evaluating models in vascular imaging applications.

\textbf{IXI (Gender):} With 561 MRI volumes labeled by gender, this dataset allows for assessing models in classifying demographic information from brain imaging data.

The 3D image dimensions vary across datasets, reflecting the anatomical and modality-specific characteristics. The pixel ranges also differ, with some datasets containing negative values due to specific preprocessing steps, such as Hounsfield unit scaling in CT images. This variation necessitates adaptable preprocessing pipelines, ensuring models can generalize across different data normalization schemes.

\subsection{Overall Dataset Diversity and Evaluation Suitability}

The selected 2D and 3D datasets collectively offer a comprehensive landscape for fine-tuning evaluation. The diversity in image dimensions accommodates various model architectures, including those tailored for handling 3D convolutions and volumetric data. The variation in pixel ranges necessitates adaptable preprocessing pipelines, ensuring that models can generalize across different data normalization schemes.

Moreover, the range of label types—from binary to multi-class—provides a spectrum of classification challenges, enabling the assessment of models' versatility and robustness. The number of samples across datasets also varies, allowing for the examination of model performance in scenarios with ample data as well as data-scarce environments.

In summary, these datasets are meticulously chosen to cover a wide array of medical imaging tasks, image modalities, and classification complexities. This diversity ensures that the fine-tuning evaluation is both thorough and reflective of real-world applications, ultimately contributing to the development of more generalized and effective machine learning models in the medical domain.}

\bibliographystyle{IEEEtran}
\bibliography{strings}

\end{document}